# Blind Underwater Image Restoration using Co-Operational Regressor Networks

Ozer Can Devecioglu, Serkan Kiranyaz, Turker Ince, and Moncef Gabbouj, *Fellow, IEEE*

*Abstract*—The exploration of underwater environments is essential for applications such as biological research, archaeology, and infrastructure maintenance. However, underwater imaging is challenging due to the water's unique properties, including scattering, absorption, color distortion, and reduced visibility. To address such visual degradations, a variety of approaches have been proposed covering from basic signal processing methods to deep learning models; however, none of them has proven to be consistently successful. In this paper, we propose a novel machine learning model, Co-Operational Regressor Networks (CoRe-Nets), designed to achieve the best possible underwater image restoration. A CoRe-Net consists of two "co-operating" networks: the Apprentice Regressor (AR), responsible for image transformation, and the Master Regressor (MR), which evaluates the Peak Signal-to-Noise Ratio (PSNR) of the images generated by the AR and feeds it back to AR. CoRe-Nets are built on Self-Organized Operational Neural Networks (Self-ONNs), which offer a superior learning capability by modulating nonlinearity in kernel transformations. The effectiveness of the proposed model is demonstrated on the benchmark Large Scale Underwater Image (LSUI) dataset. Leveraging the joint learning capabilities of the two cooperating networks, the proposed model achieves the state-of-art restoration performance with significantly reduced computational complexity and often presents such results that can even surpass the visual quality of the ground truth with a 2-pass application. Our results and the optimized PyTorch implementation of the proposed approach are now publicly shared on GitHub.

## I. INTRODUCTION

The exploration of underwater environments is crucial for various applications, including biological and archaeological research, wreckage exploration, and the maintenance of underwater pipelines and cables. The increasing use of Autonomous Underwater Vehicles (AUVs) and Remotely Operated Vehicles (ROVs) has significantly enhanced the ability to capture images in previously inaccessible areas beneath the sea. However, it is still not convenient to capture clear images with natural colors under the sea due to the varying properties of water. Compared to air, light scatters and absorbs more quickly underwater, which leads to color distortion, blurring, noise, and reduced vision and contrast. As light moves deeper into the water, shorter wavelengths such as red and yellow are quickly absorbed, leaving the water mostly blue-green in color. Further scattering of light by organic compounds and water particles reduces contrast and creates haze.

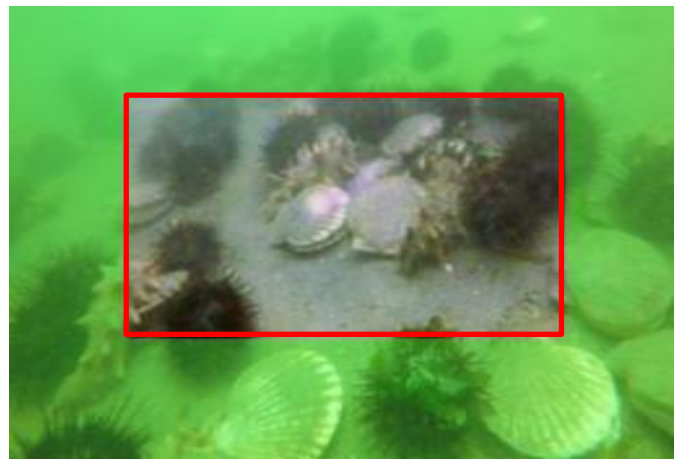

Figure 1. Sample corrupted image from the LSUI Dataset with the region inside the rectangle restored using the proposed CoRe-Net approach.

Figure 1 shows a sample underwater image from the LSUI Dataset [1]. The area inside the red rectangle shows the section that was restored using the proposed CoRe-Net network, while the area outside the red rectangle depicts the original underwater image. Comparing the restored part and the original image yields that the underwater image is affected by various artifacts, including severe noise and blurring, green saturation, reduced contrast, and visibility. A robust restoration of these corrupted underwater images has always been challenging for engineering applications since so many artifacts need to be recovered in order to convert the corrupted underwater image into a clean output.

Several prior works exist that attempted to restore underwater images. The exploration of such a wide range of approaches highlights the challenges and significance of this

O. Devecioglu and M. Gabbouj are with the Department of Computing Science, Tampere University, Tampere, Finland (e-mail: ozer.devecioglu@tuni.fi, moncef.gabbouj@tuni.fi).

S. Kiranyaz is with the Electrical Engineering Department, Qatar University, Doha, Qatar (e-mail: mkiranyaz@qu.edu.qa).

T.Ince is with the Media Engineering and Technology Department, German International University, Berlin, Germany (e-mail:turker.ince@giu-berlin.de)

problem. Signal processing-based solutions cannot perform sufficient restoration of the underwater images corrupted by a varying blend of artifacts. Deep learning models such as transformers can achieve a superior restoration performance; however, they are computationally demanding and require large datasets for training, and makes a real-time restoration infeasible.

Chiang *et al*. [2] proposed a model for wavelength compensation and image dehazing. They present a method to enhance underwater images using dehazing, attenuation correction, and artificial light compensation. It segments the scene, estimates depth, and adjusts colors to restore balance and clarity. They tested their method over a small dataset extracted from a YouTube record from Bubble Vision Company [3] and achieved 19.72 dB and 18.41 dB PSNR values in depths of 5 and 15 meters, respectively. Liu *et al*.[4] proposed an underwater image enhancement method using a deep learning network, ResNet regressor model [5]. They trained the network using paired data synthesized using Cycle GANs with real-life images. They obtained a 43.9 dB PSNR test score over the synthetic data and they obtained a 5.2 underwater image quality measure score over 221 underwater images. Luo *et al*.[6], proposed a model for underwater image restoration, focusing on color balance, contrast optimization, and histogram stretching. The model renews the R, G, and B channel values to mitigate color shifts and employs an optimized contrast technique for better transmittance. They achieved 12 dB mean PSNR in 18 image testing set. Awan *et al*.[7], propose UW-Net, that leverages discrete wavelet transform (DWT) and inverse discrete wavelet transform (IDWT) for effective feature extraction for underwater image restoration. The model was tested over LSUI and Enhancing Underwater Visual Perception (EUVP) datasets and obtained 19.87 and 19.90 dB PSNR, respectively. Although such deep network models have provided an improved and generalized restoration performance, they failed to pass the 20 dB PSNR barrier, and most of them were tested over a limited image dataset.

The current state-of-the-art performance level has been achieved by Peng et al. in [1] who introduced a U-Shape Transformers-based specialized module for color and spatial feature enhancement models to restore underwater images. They trained and tested over the LSUI dataset, which is the largest benchmark dataset ever composed, and obtained a 24.16 dB PSNR on the test set at the expense of increased complexity.

As important issue in the LSUI dataset is that there are no actual ground truth (GT) images; instead, they are manually selected and fine-tuned among the outputs of the existing restoration methods. Such a manual and subjective process obviously cannot guarantee the removal of all artifacts, especially the color variations, and certain artifacts may still persist. Another major issue is that such deep networks may overfit to such problematic GT images rather than learning the actual restoration. Such limitations highlight the need for optimization and further research to improve their efficiency and robustness in underwater environments.

In order to address the aforementioned issues and drawbacks, in this paper, we propose a novel machine learning algorithm, Co-Operational Regressor Networks (CoRe-Net) as an alternative model for Generative Adversarial Networks (GANs) [8] and its variants. GANs are distinct models with two competing networks that *compete* with each other to boost learning performance. One of the adversarial networks, the Generator, performs the primary task of image-to-image transformation. Meanwhile, the other network, the Discriminator, classifies the images as fake (produced by the generator) or real (the GT). The goal is to create such outputs that are very similar to the target GT images so that the discriminator is deceived into classifying the generated images as *real*. GANs are known to suffer from training stability problems, hence several modifications have been proposed to improve their performance [9]. In contrast, the CoRe-Net model consists of two *co-operating* networks to improve each other's performance. They are called Apprentice Regressor (AR) and Master Regressor (MR) respectively. The AR acts as a Generator by performing an image-to-image transformation. On the other hand, MR jointly learns to regress (evaluates) the actual performance level of the AR-generated images, feeds its evaluation back to the AR during training, and thus enforces AR to generate better outputs. Assuming, e.g., PSNR is the primary evaluation metric, it can directly be embedded into the CoRe-Net's loss function and the communication between AR and MR defines our algorithm as a "PSNR Boosting" by minimizing the loss (negative PSNR) function. For both networks in a CoRe-Net, Self-Organized Operational Neural Networks (Self-ONNs) [10]- [24] are used to further improve the overall learning performance. They are highly heterogeneous networks that can be characterized as the superset of CNNs with a modifiable parameter, Q, defining each kernel transformation's degree of nonlinearity (i.e., the degree of the kernel polynomials). When Q=1 for each generative neuron, the Self-ONN naturally reduces back into a conventional CNN. The nonlinear nodal operator's "on-the-fly" creation allows the network to provide the best basis functions for attaining a superior learning performance. This pilot study shows that such a highly cooperative model can outperform deep regressor networks, operational GANs and recent Transformers [1] as the prior state-of-the-art model for underwater image restoration.

The rest of this article is organized as follows: The proposed CoRe-Net model is presented in Section II with a brief explanation of Self-ONNs. In Section III, the benchmark LSUI dataset is introduced first and then the performance of the proposed model over the LSUI dataset is evaluated. Conclusions and topics for future research are presented in Section IV.

## II. METHODOLOGY

In this section, we first provide a brief overview of Self-ONNs and their key characteristics. Next, we present the Novel Image-to-Image transformation CoRe-Net model.

### A. *Self-Organized Operational Neural Networks*

The nodal operator of each generative neuron in a Self-ONN, as opposed to the convolution operator of CNNs, is capable of carrying out any nonlinear transformation that can be represented using the Taylor approximation near origin:

$$\psi(x) = \sum_{n=0}^{\infty} \frac{\psi^{(n)}(0)}{n!} x^n \qquad (1)$$

which can approximate any arbitrary function ψ well near 0. When the activation function bounds the neuron's input feature maps in the vicinity of 0, e.g., hyperbolic tangent (tanh) function, the formulation of (1) can be exploited to form a composite nodal operator. If we denote $\frac{\psi^{(n)}(0)}{n!}$ as $w_n$, the nodal operator can be expressed as follows:

$$\psi(w, x) = w_0 + w_1 x^1 + w_2 x^2 + \cdots + w_Q x^Q \qquad (2)$$

where $w_0$ is the bias which can be omitted, and $w_1$ to $w_Q$ are the Taylor series coefficients, which will be optimized during the backpropagation training.

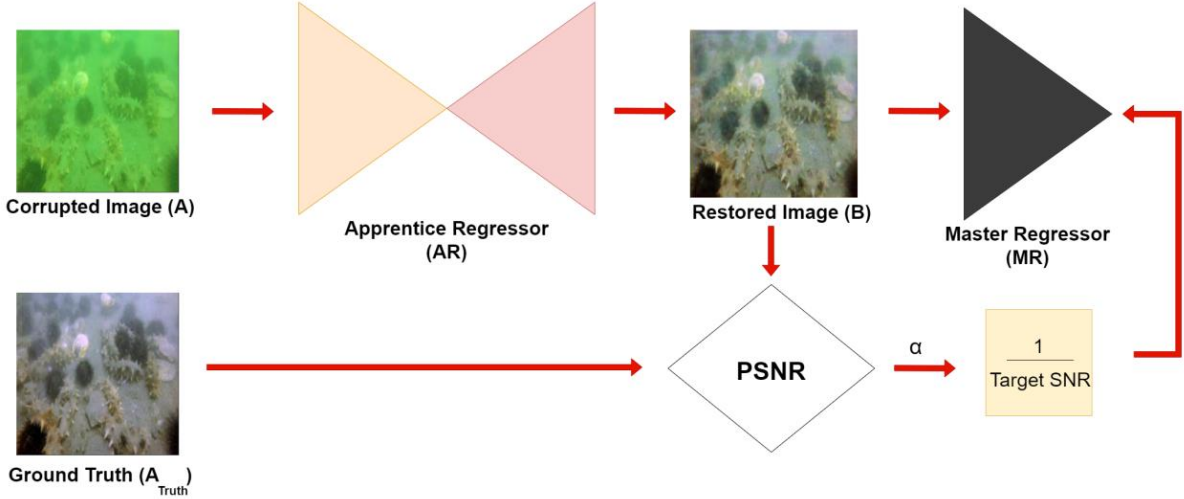

Figure 2 The general framework for the CoRe-Net framework for underwater image restoration.

Further details regarding Self-ONNs' theory and forward propagation formulations can be found in [11].

### B. Co-Operational Regressor Networks

The general framework for the novel CoRe-Net framework for underwater image restoration is given in Fig 2. The proposed CoRe-Net model consists of two ML models which are the Apprentice Regressor (AR) and Master Regressor (MR). The main target of AR is to generate high-quality restored images from corrupted counterparts. On the other hand, MR forces to boost the generation quality of the AR by learning to regress the image quality in terms of PSNR.

To train the proposed CoRe-Net model, all R, G, B input channels are first normalized between -1 and 1:

$$X_N(i) = \frac{2(X(i) - X_{\min})}{X_{max} - X_{min}} - 1 \qquad (3)$$

where $X(i)$ is the $i$th original image amplitude in the image, $X_N(i)$ is the $i$th sample amplitude of the normalized segment, $X_{min}$ and $X_{max}$ are the minimum and maximum amplitudes within the image, respectively. This will scale the image linearly in the range of [-1 1], where $X_{min} \to -1$ and $X_{max} \to 1$. Accordingly, $\max(X_i) = 1 - (-1) = 2$.

$$PSNR(X, Y) = 10 \log_{10}\left(\frac{(\max(X_i))^2}{MSE(X_i, Y_i)}\right) \qquad (4)$$

$$\alpha = PSNR(B, A_{Truth}) \qquad (5)$$

$$L_{MR} = \left\| MR(B), \frac{\alpha}{Target\ PSNR} \right\|_{L1} + \| MR(A), 1 \|_{L1} \qquad (6)$$

The main objective function of the MR is expressed in Eq. (6), where $A$ is the input image, $B$ is the restored image, $A_{Truth}$ is the corresponding ground truth, $\alpha$ is the PSNR value of the restored image and the $Target\ PSNR$ is a user-defined targeted PSNR value. The idea of the MR is to regress the PSNR value of the output and ground truth images that become the MR's input during backpropagation training. To train MR, for generated images $B$, the L1 loss between $MR(B)$ and normalized PSNR of B with respect to the target PSNR value is calculated. The PSNR value of GT images is assumed as the $Target\ PSNR$. Therefore, for $A_{Truth}$, the L1 loss between $MR(A)$ and $Target\ PSNR/Target\ PSNR = 1$ is calculated. The main objective function of the AR is expressed as,

$$L_{AR} = \varepsilon \| MR(B), 1 \|_{L1} - \beta PSNR(B, A_{Truth}) - \varphi FFL(B, A_{Truth}) \qquad (7)$$

where the fixed coefficients $\varepsilon$, $\beta$, and $\varphi$ are empirically set. Since the main objective is to generate the restored outputs with the highest possible PSNR values, the AR consists of a summation of the three different loss terms each of which contributes to this objective. First, the L1 loss between MR(B) and 1 is calculated as the feedback from the MR. Consequently, the AR will learn to generate such outputs (B) which will be evaluated by the MR and the quality difference from the GT image will be penalized. Then the actual PSNR values between $B$ and $A_{Truth}$ are calculated and penalized. Finally, the focal frequency loss (FFL) [25] is calculated between B and $A_{Truth}$. The main idea of FFL is to align patterns in the frequency domain, optimizing image restoration by focusing on major spectral components.

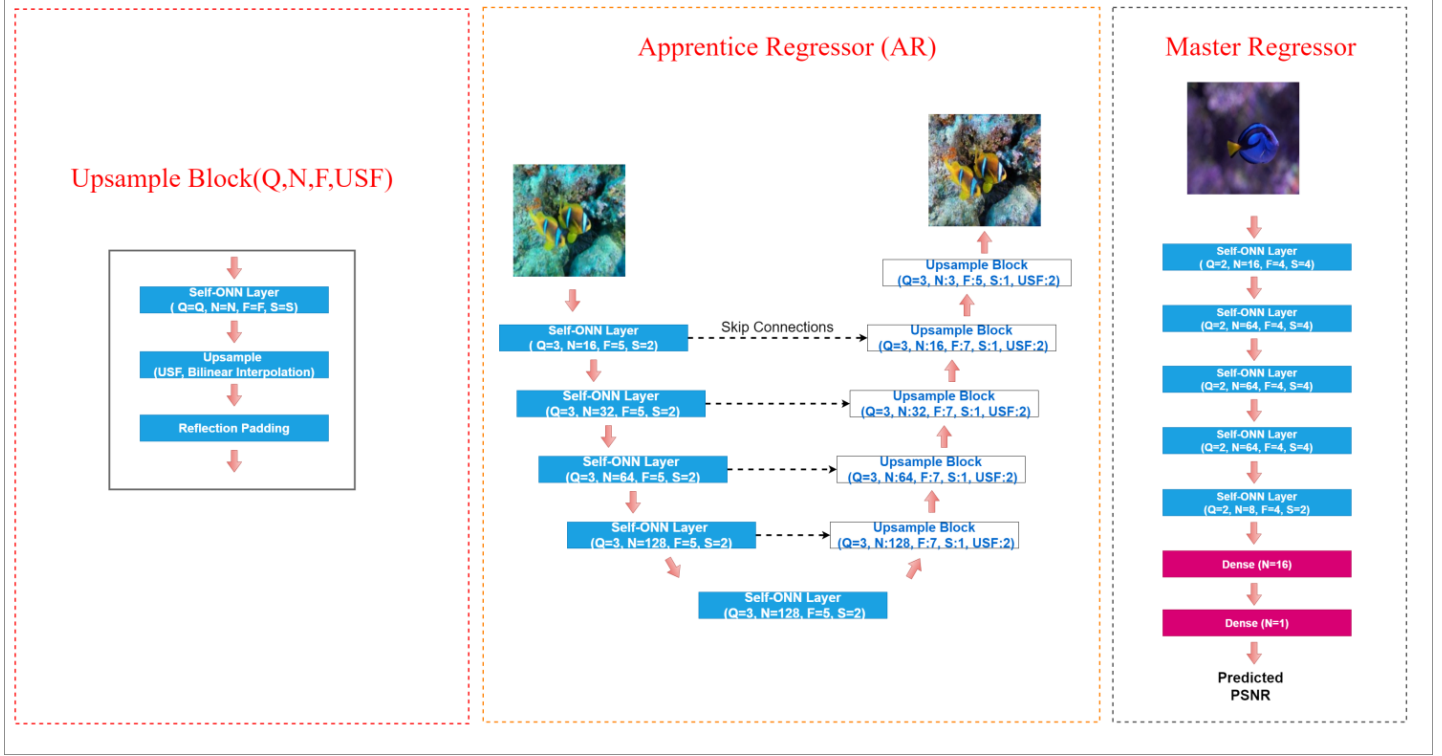

Figure 3 The Master and Apprentice Regressor architectures of CoRe-Net.

## III. EXPERIMENTAL RESULTS

This section will first introduce the benchmark, Large Scale Underwater Image (LSUI) dataset that was used in this study. Then, the experimental setup for evaluating the proposed CoRe-Net will be discussed. In Section III.C, quantitative and qualitative evaluations of the results and discussions, especially for the real use-case scenarios, are provided. In Section III.D, the second-pass CoRe-Net results are presented. Finally, in Part III.E, the computational complexity of the proposed approach is examined in-depth.

### A. Large Scale Underwater Image Dataset

LSUI is a large-scale underwater image dataset including 4279 image pairs, which covers abundant underwater scenes with high-quality reference images. The dataset also contains real underwater images from other benchmark underwater datasets. The ground-truth (GT) set of the underwater images are formed using 2-step evaluations. First, they used an ensemble model that used 18 existing underwater image enhancement methods [1]. After the first round, volunteers and experts voted for the best-enhanced image to form the paired dataset.

### B. Experimental Setup

The Self-ONN models for AR and MR are shown in Figure 3. For the AR model, a 10-layer U-Net model is used with 5 operational layers and 5 upsampling (by 2) and operational layers with residual connections (instead of transposed convolution layers). For the decoder side of the AR, kernel sizes are all set as 5. The stride is set as 2 for all down-sampling operational layers. For the encoder side of the AR, all kernel sizes are set as 7. $Q$ values for all layers are set to 3. On the other hand, the MR model consists of 2 dense layers and 5 operational layers with a kernel size of 4. The strides for operational layers are set as 4, 4, 4, 2, and 2, respectively. $Q$ values for operational layers are set to 2. For all experiments, the batch size is set to 1 and a maximum of 5000 Back-Propagation (BP) iterations are used. The Adam optimizer with an initial learning rate of $10^{-5}$ is used via BP. The $\varepsilon$, $\beta$, and $\varphi$ in Eq. are empirically set as 3, 0.05 and 100, respectively. The target PSNR is set to 40 dB. We implemented both AR and MR networks using the FastONN library [13] based on PyTorch. The benchmark dataset, our results, and the optimized PyTorch implementation of the proposed approach are now publicly shared with the research community in [26]. The LSUI dataset is partitioned into train and test sets as 90 % and 10 % respectively as in [1] for fair comparative evaluations.

To compare the results of the CoRe-Net, the equivalent operational (Op.) U-Net and GAN with the identical Self-ONN models are also implemented. Op. GANs [21] are the recent variants of the conventional GANs with a superior learning performance. The conventional objective functions for Op. GANs as expressed in Eq. (8) are used as the loss functions during training.

$$\min_{OG} \max_{OD} L_{OP-GAN}(OG, OD) \\ = E[\log(OD(GT))] \\ + E[\log(1 - OD(OG(X)))] \quad (8)$$

where, $OG$ and $OD$ represent the Op. Generator and Discriminator, respectively. $X$ denotes the input, and $GT$ stands for the ground truth.

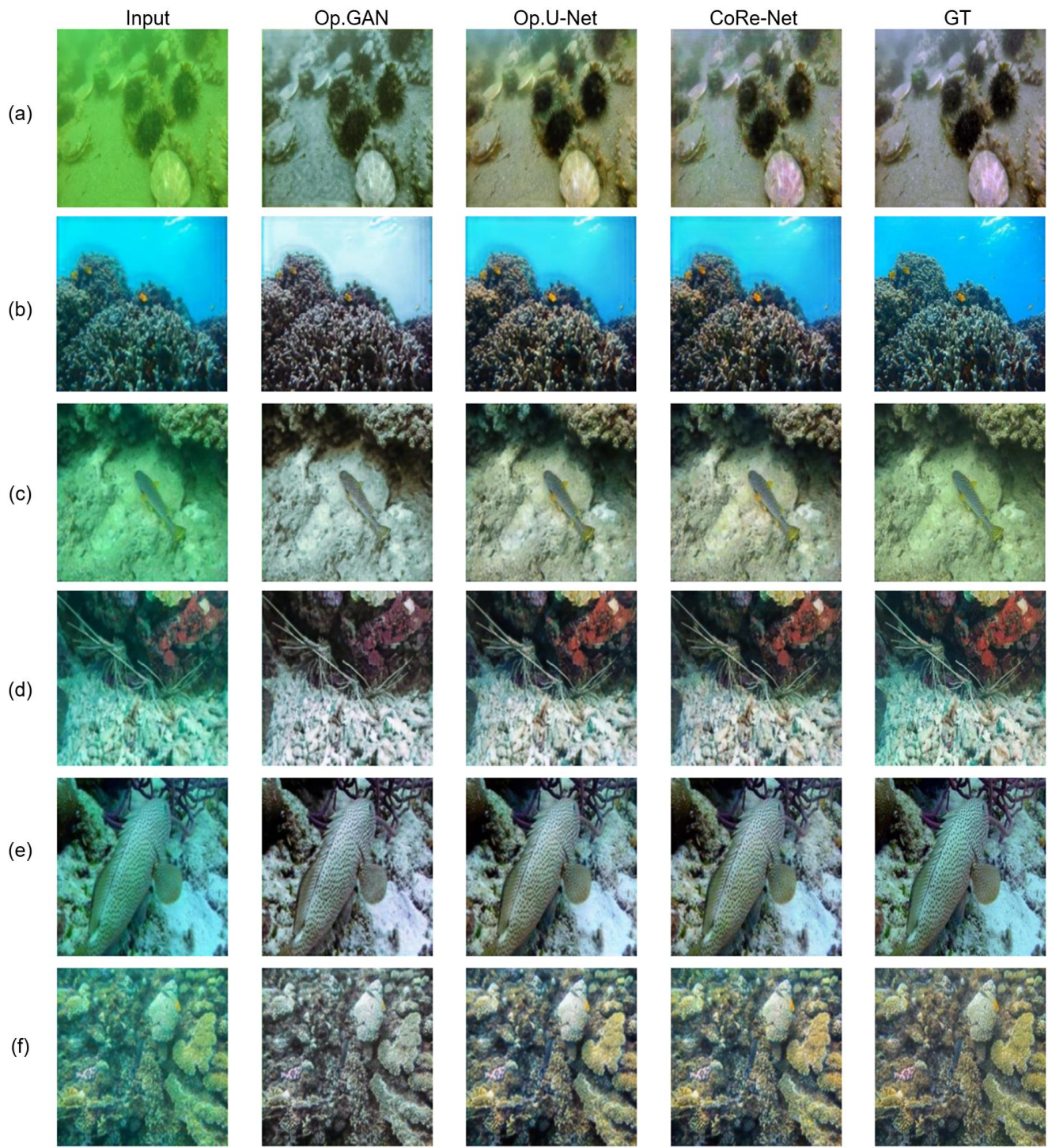

Figure 4 Sample underwater image restoration results from Op. GAN [21], Op. U-Net [22], CoRe-Net (proposed), and the corresponding Ground Truth (GT).

## C. Results

This section presents both quantitative and qualitative (visual) evaluations of the proposed CoRe-Net model against the corresponding single regressor (Op. U-Net), Op. GAN, and the recent competing models [1], [27]-[31]. Table 1 shows quantitative evaluations in terms of PSNR score along with the model complexities. For a fair comparison, both Op. U-Net [22] and the Generator of the Op. GAN [21] has the same AR model for restoration. Moreover, the training and hyper-parameters are also kept identical.

**Table 1** The underwater image restoration performance over LSUI Dataset.

|  | PSNR (dB) | Parameters (M) |
|---|---|---|
| Op. U-Net [22] | 23.62 | 7.2 |
| Op. GANs [21] | 21.08 | 7.2 |
| **CoRe-Net** | **24.54** | **7.2** |
| U-Shape Transformer [1] | 24.16 | 65.6 |
| UColor [27] | 22.91 | 157.4 |
| UIE-DAL [28] | 17.45 | 18.82 |
| UGAN [29] | 19.79 | 57.17 |
| FunIE [30] | 19.79 | 7.01 |
| WaterNet [31] | 17.73 | 24.81 |

From the table, we can clearly state that the cooperating networks MR and AR in the CoRe-Net significantly improve restoration performance compared to the Op. U-Net and GAN's performances. The second part of the evaluation contains a comparison of our proposed model with respect to recent competing methods including the state-of-the-art method (SoTa), [1]. The CoRe-Net achieves the highest PSNR of 24.54 dB, indicating superior image quality with a significantly lower complexity compared to the Transformer in [1]. In contrast, CoRe-Net has significantly surpassed (around 4dB in PSNR) the only model with comparable complexity, FunIE [30].

For qualitative evaluation, some random underwater image restoration results from the operational GAN, U-NET, and CoRe-Net, along with the corresponding Ground Truth (GT) models are shown in Figure 4. The rows (labeled a–f) contain different underwater scenes, each with the restoration outputs demonstrating the performance of the models. Overall, the real underwater image input has an overlaying green or blue color and suffers from low contrast and various artifacts. Results from Operational GAN-based restoration performed more color-corrected outputs but tended to have a subtle tint and lower sharpness compared to the ground truth. Results from Operational U-Net can offer certain improvements over GAN, with sharper images and better contrast. However, especially the object colors are still slightly different from the ground truth. The proposed CoRe-Net network provides a sharper, cleaner output with the true object(s) and background colors. The color correction is indeed closer to the GT scene, with the improved contrast and details.

## D. The 2-pass Restoration Approach

In this approach, two consecutive restoration passes are applied. Quantitatively, the second-pass CoRe-Net restoration achieves a mean PSNR value of 20.57 dB across the entire test set. This PSNR score is approximately 4 dB lower than the state-of-the-art score obtained during the first pass of CoRe-Net. However, despite the lower PSNR, the visual quality of the restored images appears to improve further as some sample examples are shown in Figure 6. More 2-pass results are presented Figure 7 and Figure 8 in the Appendix A and also shared on GitHub [26].

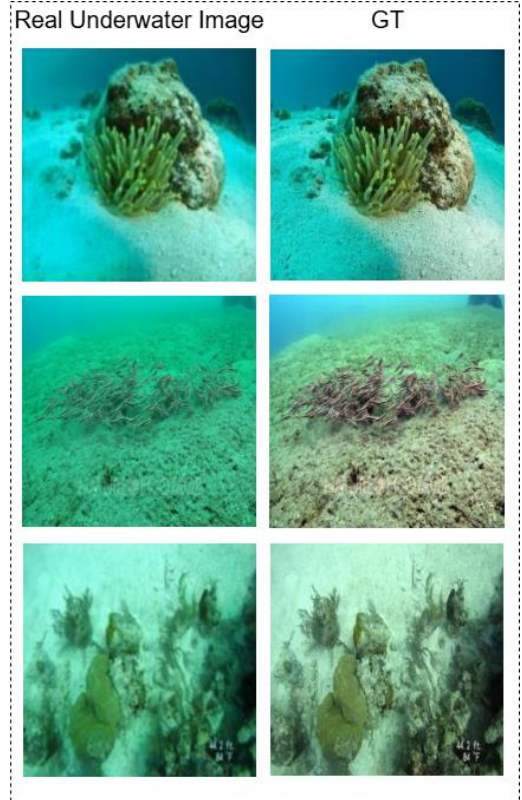

**Figure 5.** Sample input images and their corresponding GT images from the LSUI dataset.

All results clearly show that the second-pass CoRe-Net restoration results are substantially better color-corrected and usually sharper than both the first pass and also the corresponding GT images. Despite this visual improvement, the second pass outputs have naturally led to a decrease in the PSNR values due to the increased visual difference between the corresponding GT images.

This highlights one of the primary challenges in underwater image restoration which is the lack of true GT images. As mentioned earlier, this benchmark dataset uses the manually selected GT images among a limited set of results. Therefore, the GT images are only as good as the selected model's restoration performance can result. Some GT images with apparent color deficiencies are shown in Figure 5 where GT images have slightly reduced the overlaid green tones. Obviously, there is still room for further restoration and in particular, the removal of the overlaid green tones, which is actually achieved in the Core-Net's second pass results as shown in Figure 6 - Figure 8.

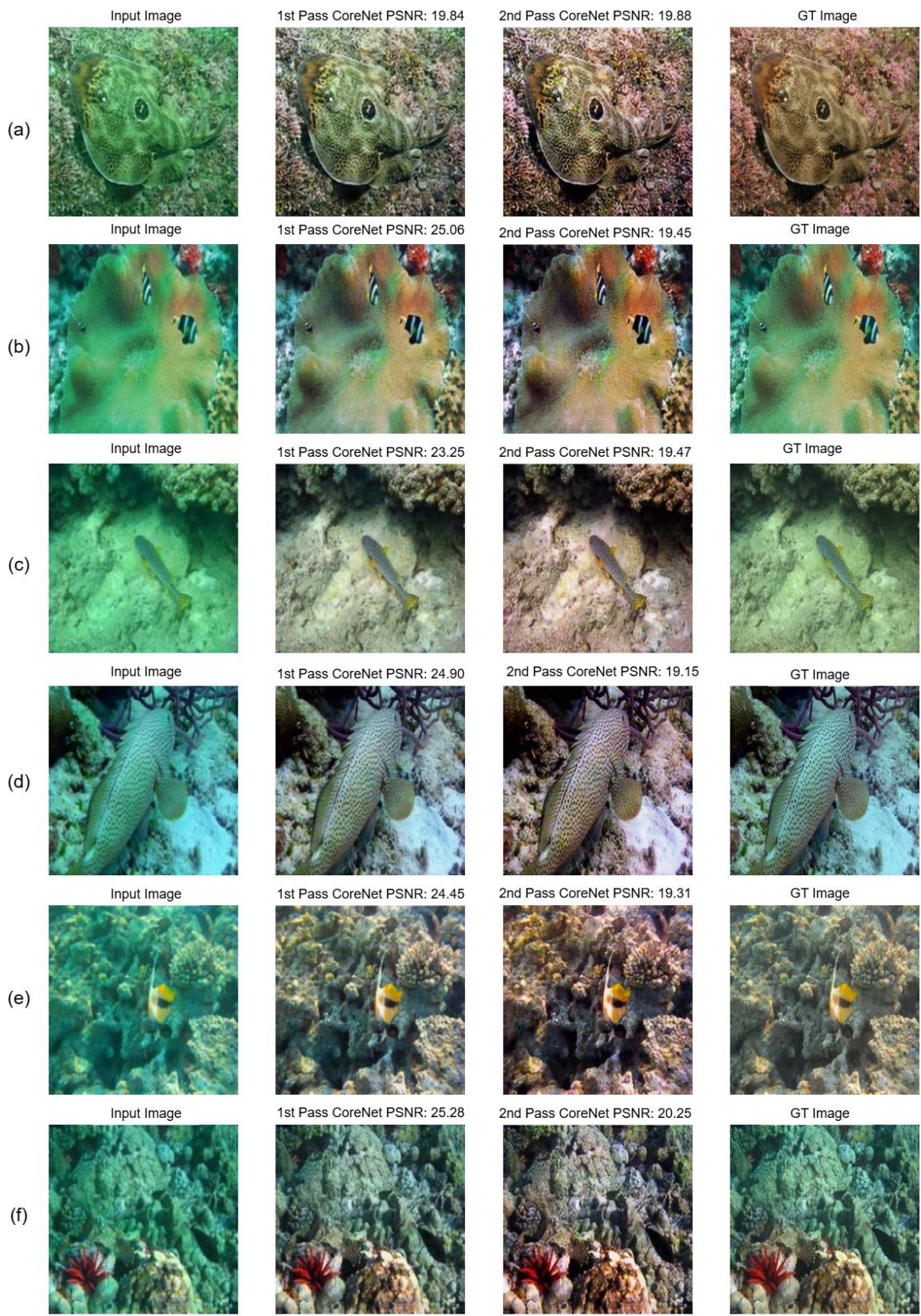

**Figure 6** A visual comparison of the underwater image restoration results from the first and second pass of the CoRe-Net, and the corresponding Ground Truth (GT).

*E. Computational Complexity Analysis*

For the computational complexity analysis, the network size, total number of parameters (PARs), and inference time for each network configuration are presented in this section. Detailed PAR formulations for Self-ONNs are available in [13]. A 2.2 GHz Intel Core i7 computer with 16 GB of RAM and an NVIDIA GeForce RTX 3080 graphics card was used for all experiments. The proposed network has a total of 7.2 M parameters. The processing time to restore a 256x256x3 image takes around 6.1 msec for a single CPU implementation. This shows that the proposed transformation can achieve approximately 150 times faster than the real-time requirements with a single CPU, indicating the potential for real-time implementation even on low-cost, low-power hardware.

## IV. CONCLUSIONS

The restoration of the underwater images is a challenging task due to the varying set of distortions caused by light scattering, absorption, complex underwater environment, and the lack of real GT images. Although traditional signal processing techniques have been utilized for underwater image enhancement, they often fall short of addressing the diverse and severe degradations present in underwater image acquisition systems. Recent advancements in deep learning, including transformer-based models can outperform prior approaches; however, they frequently face obstacles related to high computational demands, large dataset requirements, and challenges in generalizing to varied underwater conditions.

The proposed CoRe-Net model introduces a cooperative learning framework between the AR and MR networks to boost learning performance. Unlike adversarial learning in traditional GANs, CoRe-Net's cooperative learning model ensures that AR focuses on generating high-quality image transformations, while MR continuously guides and enforces the enhancement of PSNR values with a direct feedback into its loss function. The blind restoration results reported in this study show that this cooperative learning approach enables more efficient training without increasing the model depths and complexity. Overall, the proposed CoRe-Net model achieves the new state-of-the-art restoration performance in the first-pass restoration outputs. Particularly with the second pass, we report that CoRe-Net can further improve the restoration performance often beyond the quality level of the GT images, which is bound to the restoration method manually selected by a group of volunteers. As a result, the 2-pass CoRe-Net results can now be used to further improve the GT images in this benchmark dataset.

The significant and novel contributions of this study can be summarized as follows:
- This study presents CoRe-Net, a novel image-to-image transformation framework.
- This novel study demonstrates state of art restoration performance over the benchmark LSUI dataset.
- Due to the compact computational complexity of the CoRe-Net AR model, the proposed approach can be implemented in real-time even on low-cost, low-power sensorial hardware.
- The 2-pass CoRe-Net approach can be used to improve the GT images in the LSUI dataset.

Our future research will explore the CoRe-Net's applicability in other similar imaging problems such as images taken during the night or in poor visibility (e.g., severe misty environment). Enabling AR to multiple outputs and engaging the MR to select the best possible output among them will be another promising research direction, which can further boost restoration performance.

APPENDIX A

In this Appendix, we shall provide more 2-pass CoRe-Net results along with the 1$^{st}$ pass and the GT images as shown in Figure 7 and Figure 8. More results can be seen in the shared GitHub repository in [26].

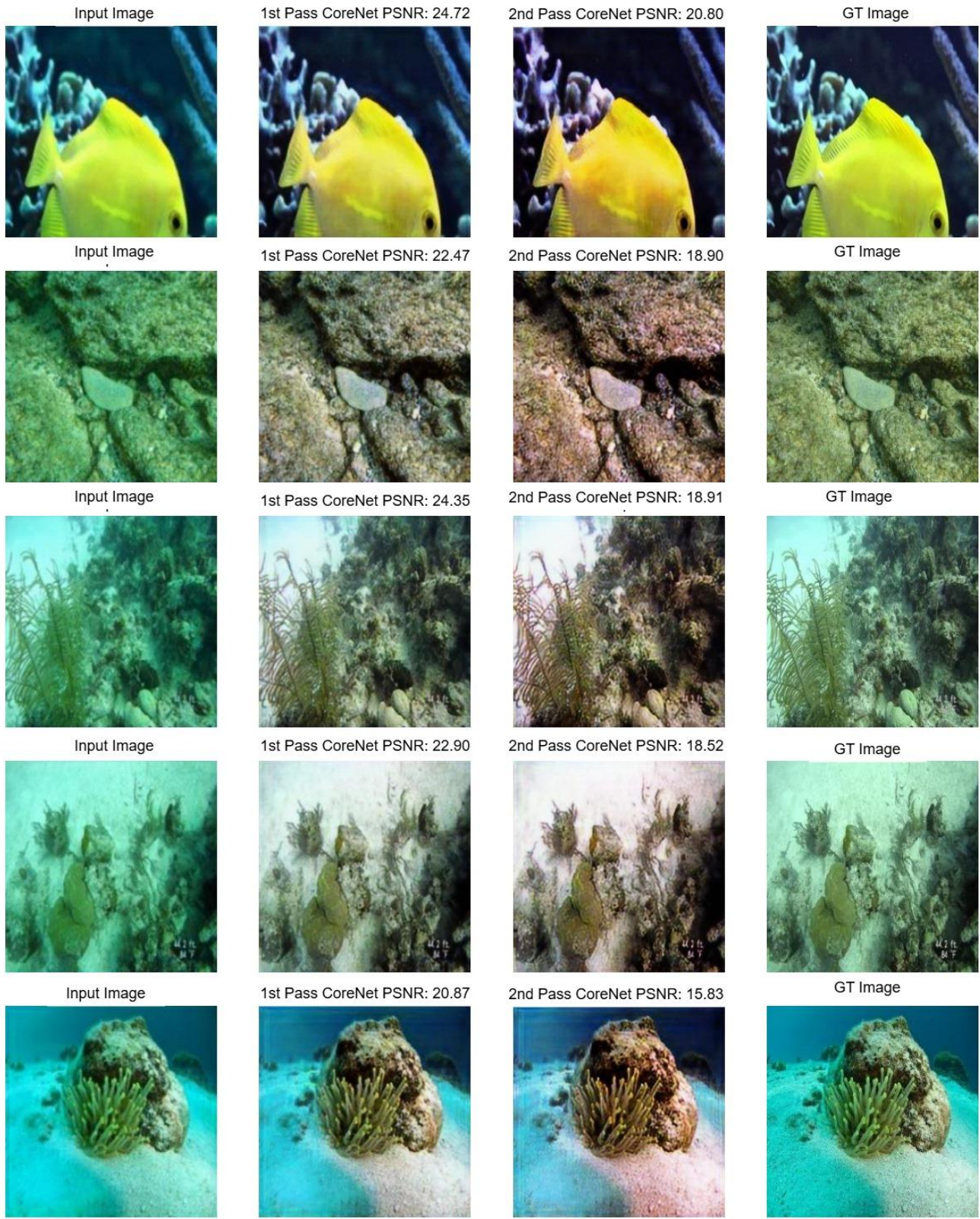

**Figure 7** A visual comparison of the underwater image restoration results from the first and second pass of the CoRe-Net, and the corresponding Ground Truth (GT).

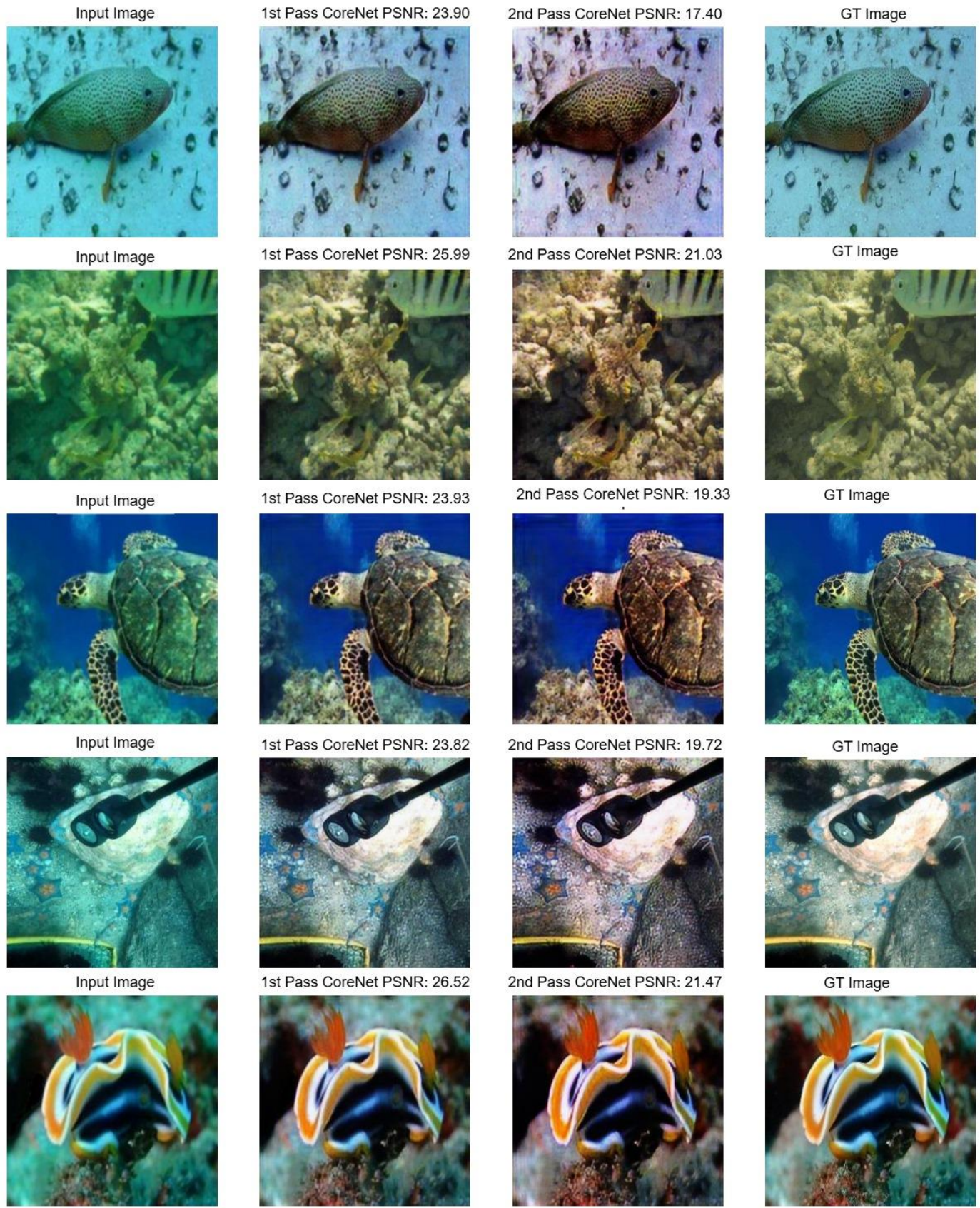

**Figure 8** A visual comparison of the underwater image restoration results from the first and second pass of the CoRe-Net, and the corresponding Ground Truth (GT)